\documentclass[11pt, logo, onecolumn, copyright, colorlinks=true, allcolors=blue]{nvidiatechreport}
\usepackage{graphicx} 
\usepackage{subcaption,ragged2e}
\usepackage[round]{natbib}

\usepackage[normalem]{ulem}
\usepackage[utf8]{inputenc} 
\usepackage[T1]{fontenc}    
\usepackage{url}
\usepackage{tikz}
\usepackage{enumitem}
\usetikzlibrary{positioning}
\usetikzlibrary{calc}
\usepackage{array}
\usepackage{booktabs}
\usepackage{wrapfig}
\usepackage{caption} 
\usepackage{listings}
\usepackage{adjustbox}
\usepackage{footnote}
\usepackage{multirow}
\usepackage{amsmath}
\usepackage{amsfonts}
\usepackage{breakcites}
\usepackage{blindtext}
\usepackage{svg}
\usepackage[most]{tcolorbox}
\usepackage{tabularx}
\usepackage{graphicx} 
\usepackage{xcolor}
\usepackage{colortbl}
\newcommand{\ourmodel}{Nemotron 3 Nano\xspace}
\newcommand{\ourbasemodel}{Nemotron-3-Nano-30B-A3B-Base\xspace}

\newcommand{\remove}[1]{}

\title{NVIDIA Nemotron 3: Efficient and Open Intelligence}
\author{\large NVIDIA}
\date{}

\begin{document}

\begin{abstract}
\large \textbf{Abstract}
\normalsize

We introduce the Nemotron 3 family of models—Nano, Super, and Ultra. These models deliver strong agentic, reasoning, and conversational capabilities. The Nemotron 3 family uses a Mixture-of-Experts hybrid Mamba–Transformer architecture to provide best-in-class throughput and context lengths of up to 1M tokens. Super and Ultra models are trained with NVFP4 and incorporate LatentMoE, a novel approach that improves model quality. The two larger models also include MTP layers for faster text generation. All Nemotron 3 models are post-trained using multi-environment reinforcement learning enabling reasoning, multi-step tool use, and support granular reasoning budget control.

Nano, the smallest model, outperforms comparable models in accuracy while remaining extremely cost-efficient for inference. Super is optimized for collaborative agents and high-volume workloads such as IT ticket automation. Ultra, the largest model, provides state-of-the-art accuracy and reasoning performance. Nano is released together with its technical report and this white paper, while Super and Ultra will follow in the coming months. We will openly release the model weights, pre- and post-training software, recipes, and all data for which we hold redistribution rights.

\end{abstract}

\maketitle

\section{Introduction}
\label{sec:intro}

We announce NVIDIA Nemotron 3, the most efficient family of open models with leading accuracy for agentic AI applications. The Nemotron 3 family of models use a Mixture-of-Experts hybrid Mamba-Transformer architecture that pushes the accuracy-to-inference-throughput frontier~(\S\ref{subsec:hybridmoe}). State-of-the-art accuracy and high inference throughput enable developers to build and scale up complex multi-agent environments. Further, the Nemotron 3 family of models support a context length of up to 1M tokens which helps accelerate tasks that require long contexts like long slices of code, large conversation histories, and extensive documents for RAG pipelines~(\S\ref{subsec:lc}). Nemotron 3 models support inference-time reasoning budget control~(S\ref{subsec:budgetcontrol}) and are trained using a diverse set of RL environments. The diverse set of environments help Nemotron 3 models achieve superior accuracy across a broad range of tasks like competitive coding, competition math, and agentic tool use~(\S\ref{subsec:multienvRL}). 

In addition to the above, Nemotron 3 Super and Ultra are trained with NVFP4~(\S\ref{subsec:nvfp4}). Super and Ultra utilize LatentMoE, a novel approach that helps gain accuracy without sacrificing inference throughput or latency~(\S\ref{subsec:latentmoe}). We also incorporate MTP layers in the two larger models to improve the efficiency of long-form text generation workloads~\citep{gloeckle2024better}. Additionally, training with MTP provides modest improvements in model quality~\citep{deepseekai2025deepseekv3technicalreport}~(\S\ref{subsec:mtp}).  

The Nemotron 3 family of models are open and transparent. We will release all the model weights, over 10 trillion tokens of datasets, and training recipes.     

In the following section, we discuss the key technologies used to build Nemotron 3.



\section{Features and Technologies}
\label{sec:keytech}

\subsection{Hybrid MoE}
\label{subsec:hybridmoe}

\begin{figure}[!t]
\centering
\includegraphics[width=0.9\linewidth]{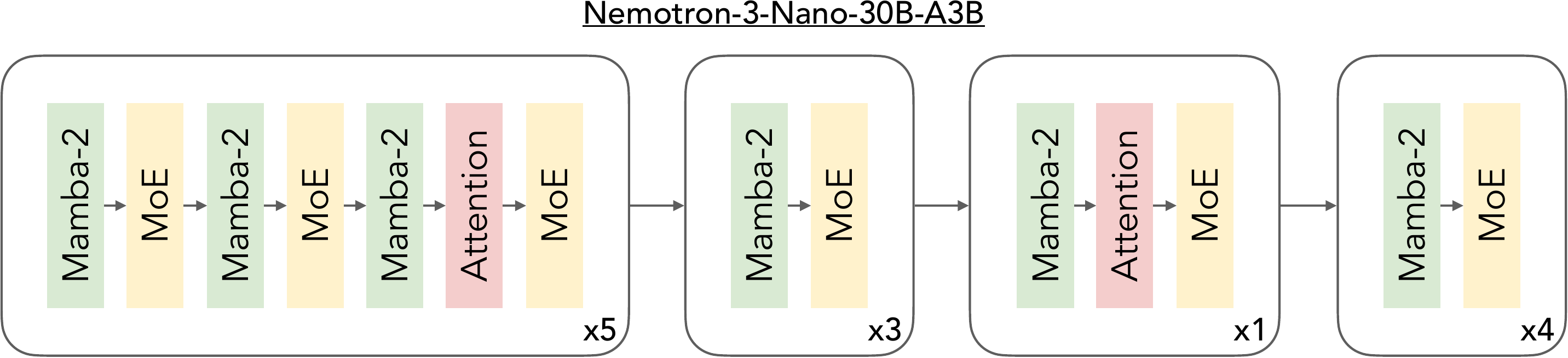}
\caption{Nemotron 3 models (e.g., Nemotron Nano 3) leverage a hybrid Mamba-Transformer MoE architecture consisting predominantly of interleaved Mamba-2 and MoE layers, with a select few self attention layers.}
\label{fig:layer-pattern}
\end{figure}

The Nemotron 3 family of models utilize a hybrid Mamba-Transformer MoE architecture. This architecture is chosen with inference efficiency in mind, particularly for reasoning workloads, but also provides better or on-par accuracy compared to standard Transformers~\citep{waleffe2024empiricalstudymambabasedlanguage, nvidia2025nemotronhfamilyaccurateefficient, basant2025nvidia}. Specifically, rather than interleaving mixture-of-expert (MoE) layers with expensive self-attention layers---which need to attend over a linearly increasing KV Cache during generation---Nemotron 3 models predominantly interleave MoE layers with cheaper Mamba-2 layers~\citep{dao2024transformersssmsgeneralizedmodels}---which require storing only a constant state during generation. Only a select few attention layers are included in Nemotron 3 models. For example, we show the layer pattern for Nemotron Nano 3 in Figure~\ref{fig:layer-pattern}.

\begin{figure}[t]
    \centering
    \includegraphics[width=\linewidth]{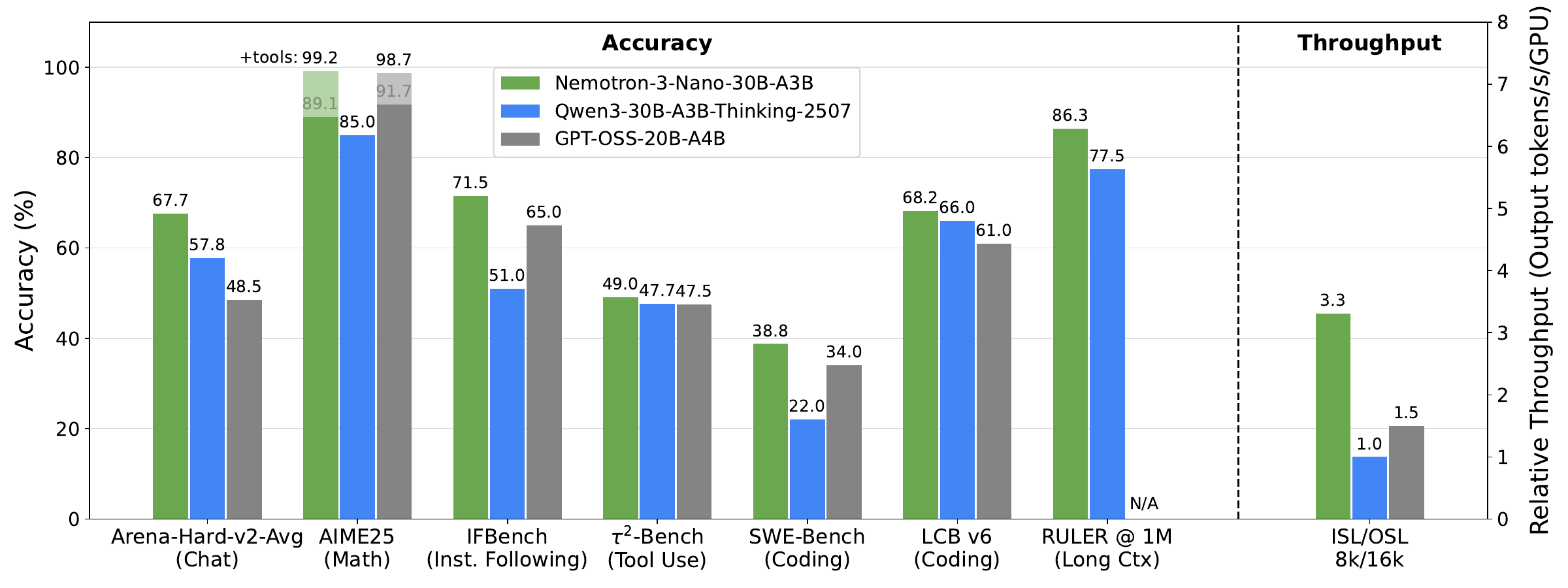}
    \caption{The hybrid Mamba-Transformer MoE architecture used by Nemotron 3 models can achieve state-of-the-art accuracy on leading reasoning benchmarks and ultra-long-context tasks while providing throughput improvements over similarly sized Transformer MoEs. For details, please see the Nemotron Nano 3 technical report. 
    }
    \label{fig:nanov3}
\end{figure}

By minimizing expensive self-attention layers, Nemotron 3 models can achieve higher inference throughput compared to similarly-sized Transformer MoEs for common reasoning workloads (e.g., 8k input sequence length / 16k output sequence length). For example, Nemotron 3 Nano 30B-A3B achieves 3.3$\times$ higher throughput compared to Qwen3-30B-A3B (Figure~\ref{fig:nanov3}), with further speedups at longer sequences. Yet the hybrid architecture can also achieve state-of-the-art accuracy, even on long-context lookup tasks (e.g., RULER on 1M token input sequences, see Figure~\ref{fig:nanov3}). 

Overall, the Nemotron 3 architecture leverages a balanced combination of mixture-of-expert layers that allow for sparse parameter scaling and lead to higher accuracy for a given compute budget, self-attention layers that enable high-fidelity all-to-all information routing, and Mamba-2 layers that enable sequence modeling with fixed inference-time computation and memory. 

\subsection{LatentMoE: Hardware-Aware Expert Design for Improved Accuracy per Byte}
\label{subsec:latentmoe}

\begin{figure}[h!tb]
    \centering
    \begin{subfigure}[b]{0.42\textwidth}
        \includegraphics[width=\textwidth]{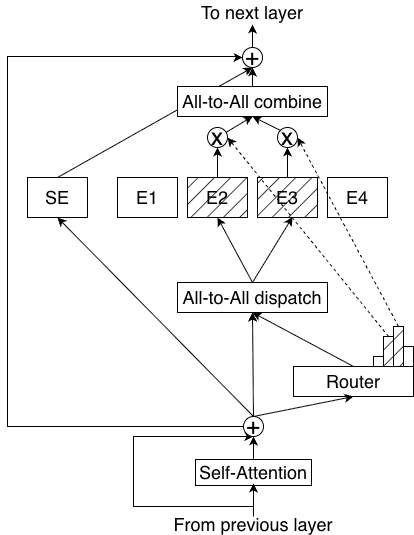}
        \caption{Standard MoE architecture.}
        \label{fig:standard_moe}
    \end{subfigure}
    \hfill
    \begin{subfigure}[b]{0.45\textwidth}
        \includegraphics[width=\textwidth]{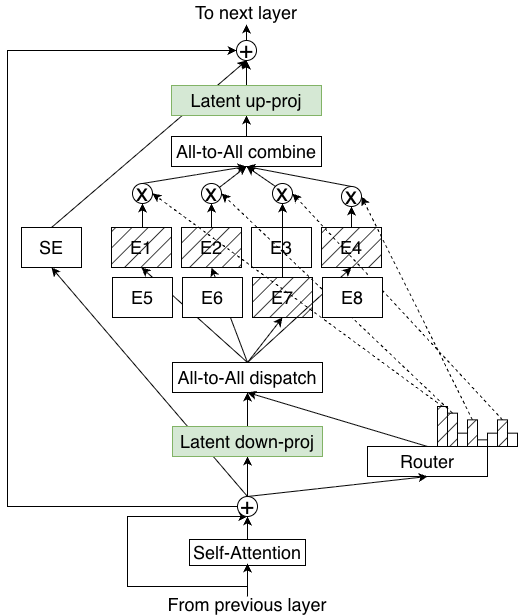}
        \caption{LatentMoE architecture.}
        \label{fig:latent_moe}
    \end{subfigure}
    \caption{Standard MoE vs. LatentMoE architectures. In LatentMoE, tokens are projected from the model hidden dimension $d$ into a smaller latent dimension $\ell$ for expert routing and computation, which reduces routed parameter loads and all‑to‑all traffic by a factor of $d / \ell$ (typically about $4\times$). We use this efficiency to increase both the total number of experts and the top-$K$ active experts per token by the same factor $d/\ell$, which improves accuracy per byte while keeping overall inference cost approximately constant.}
\end{figure}

Transformer models are typically deployed in two distinct settings: latency-focused deployments that prioritize response time, and throughput-focused deployments that maximize token processing capacity.
Mixture of Experts (MoE) layers face fundamentally different performance bottlenecks depending on the scenario.
In latency-focused deployments, the model processes tens to hundreds of tokens at a time to minimize end-to-end latency. In this regime, MoE computation is memory-bandwidth-bound: reading expert weights from memory dominates the cost and far exceeds the actual computation time. Each expert matrix has size $d \times m$, where $d$ is the model hidden dimension and $m$ is the expert FFN intermediate dimension, so reducing memory bandwidth costs requires decreasing either $d$ or $m$.
In throughput-focused deployments, the model processes thousands of tokens per iteration to maximize throughput. In this regime, the all-to-all communication required to dispatch tokens to experts and aggregate results emerges as the primary bottleneck. The communication volume scales linearly with the number of top-$K$ active experts $K$ and the hidden dimension $d$, but is independent of the expert FFN intermediate dimension $m$.
At the same time, the expressive power of FFN layers is primarily controlled by the effective nonlinear budget, which is roughly proportional to $K \times m$.

Our objective is to improve model quality without compromising inference throughput or latency.
Guided by the above insights, we adopt a specific design choice. To improve accuracy per byte, we shrink the \emph{routed} expert input dimension $d$ to reduce communication and memory costs, and reinvest
the saved capacity into increasing the nonlinear budget and expert diversity by scaling up both the
total number of experts $N$ and the top-$K$ active experts per token. LatentMoE is a novel architecture that implements this strategy.

The LatentMoE architecture is illustrated in Figure~\ref{fig:latent_moe}. Each token embedding is first projected from the original hidden dimension $d$ into a latent representation of smaller dimension $\ell < d$, routed to an expanded set of experts that operate entirely in this latent space, and then projected back to the original hidden dimension $d$.
By shifting routed expert computation and all-to-all traffic to the latent space, both per-expert weight loads and communication payloads are reduced by a factor of $d / \ell$ compared to a standard MoE. 
We use these parameter and bandwidth savings to increase both the total number of experts from $N$ to $N' = N \cdot d / \ell$ and the top-$K$ active experts per token from
$K$ to $K' = K \cdot d / \ell$. The reduction in dimensionality offsets the increase in expert count and active experts, which enables higher model quality at similar computational and communication budget.
To preserve quality, all non-routed computations, including the MoE routing gate (gating network), shared expert computation, and non-expert layers, remain in the original hidden dimension $d$, since they do not significantly contribute to the targeted bottlenecks.

\begin{table}
    \centering
    \caption{Comparison of downstream task accuracy between Standard MoE and LatentMoE. The LatentMoE architecture consistently outperforms the standard MoE baseline across all evaluated tasks.}
    \label{tab:latent_moe_8B_1T_results}
    \begin{tabular}{c||c|c|c|c|c}
        \toprule
        \multirow{2}{*}{\textbf{Model}} & \multicolumn{5}{c}{\textbf{Accuracy (\%)}} \\
        \cline{2-6}
         & \textbf{MMLU-Pro} & \textbf{MMLU} & \textbf{Code} & \textbf{Math} & \textbf{\begin{tabular}{@{}c@{}}Commonsense\\Understanding\end{tabular}} \\
        \hline
        \begin{tabular}{@{}c@{}}\textbf{Standard MoE}\\(8.09B active / 72.6B total)\end{tabular} & 48.30 & 70.10 & 51.95 & 78.32 & 81.73 \\
        \hline
         \begin{tabular}{@{}c@{}}\textbf{LatentMoE}\\(8.02B active / 72.8B total)\end{tabular} & \textbf{52.87} & \textbf{72.11} & \textbf{55.14} & \textbf{80.19} & \textbf{82.10} \\
        \bottomrule
    \end{tabular}
\end{table}

Table~\ref{tab:latent_moe_8B_1T_results} compares the downstream performance of Standard MoE and LatentMoE. To provide a comprehensive evaluation, we report aggregated scores: ``Code'' averages HumanEval, HumanEval+, MBPP, and MBPP+; ``Math'' averages GSM8K CoT and MATH-500; and ``Commonsense Understanding'' averages RACE, ARC-Challenge, HellaSwag, and Winogrande. Both models feature 8 billion active and 73 billion total parameters, and were trained for 1 trillion tokens using identical hyperparameters. Specifically, the Standard MoE model uses a hidden dimension of size $d=4096$ and 128 total experts with 6 active experts, while LatentMoE uses a latent dimension $\ell=1024$ and 512 total experts with 22 active experts. As the results demonstrate, LatentMoE consistently outperforms the Standard MoE baseline across all evaluated tasks.
\subsection{Multi-Token Prediction (MTP)}
\label{subsec:mtp}
Multi-Token Prediction (MTP) has emerged as a highly effective technique for improving both the 
\emph{accuracy} and the \emph{inference efficiency} of large language models. Prior work—
including DeepSeek-V3~\citep{deepseekai2025deepseekv3technicalreport} and the original MTP 
formulation~\citep{gloeckle2024better}—shows that predicting multiple future tokens provides richer 
training signals and encourages models to plan several steps ahead. These auxiliary predictions also 
serve naturally as draft tokens for speculative decoding~\citep{leviathan2023fast}, enabling 
substantial end-to-end acceleration without requiring a separate draft model.

In Nemotron~3, integrating MTP leads to consistent gains across validation loss and a broad range of downstream benchmarks, including general knowledge, code generation, common-sense understanding, reading comprehension, and math. On an ablation study using the 8B active parameters transformer MoE model base model, MTP improves performance by roughly 2.4\% on average across benchmarks (see Table~\ref{tab:moe_mtp_results}). These improvements reflect MTP’s ability to provide denser supervision and enhance the model’s multi-step reasoning capabilities. From a systems perspective, MTP introduces minimal additional FLOPs and integrates seamlessly into our training workflow, delivering substantial speculative-decoding benefits~\citep{gloeckle2024better} while maintaining high overall efficiency.

\begin{table}[h!]
\centering
\begin{tabular}{l|cc}
\toprule
\textbf{Task} & 
\textbf{Baseline (8B MoE Base)} &
\textbf{Baseline + MTP} \\
\midrule

\rowcolor{black!5}
\multicolumn{3}{l}{\textbf{General Knowledge}} \\
MMLU (5-shot, acc)        & 70.06 & \textbf{71.26} \\
MMLU-Pro (5-shot, CoT EM) & 45.05 & \textbf{47.84} \\
\midrule

\rowcolor{black!5}
\multicolumn{3}{l}{\textbf{Code}} \\
MBPP-Sanitized (3-shot) & 65.58 & \textbf{66.89} \\
\midrule

\rowcolor{black!5}
\multicolumn{3}{l}{\textbf{Commonsense Understanding}} \\
ARC-Challenge (25-shot, acc\_norm) & 86.43 & \textbf{88.05} \\
WinoGrande (0-shot, acc)        & 74.59 & \textbf{75.45} \\
\midrule
\rowcolor{black!5}
\multicolumn{3}{l}{\textbf{Reading Comprehension}} \\
RACE (0-shot, acc)              & 84.02 & \textbf{85.36} \\
\midrule
\rowcolor{black!5}
\multicolumn{3}{l}{\textbf{Math}} \\
GSM8K (8-shot, acc)   & 82.49 & \textbf{84.46} \\
\bottomrule
\end{tabular}

\caption{
Accuracy scores with and without MTP on a simple 8B active parameter transformer MoE base model trained on 1T tokens. We observe improvements in accuracy on multiple tasks spanning different categories.
}
\label{tab:moe_mtp_results}
\end{table}

A major practical benefit of MTP in Nemotron~3 is its strong synergy with speculative decoding. 
The predictions produced by MTP exhibit high agreement with the base model, enabling fast, 
low-latency generation — particularly beneficial in batch-size--1 and long-form generation scenarios.  
We designed a lightweight MTP module that achieves around 97\% acceptance on the first two predicted tokens in an ablation study on an 8B active MoE model. Overall, MTP enriches the training 
signal, enhances the model’s ability to anticipate multiple future steps, provides high-quality 
speculative-decoding predictions, and accelerates text generation~\citep{gloeckle2024better}. 



\subsection{NVFP4 Training}
\label{subsec:nvfp4}

We demonstrate stable and accurate pretraining on a hybrid Mamba-MoE architecture for up to 25T tokens using the NVFP4 number format. Weight, activation, and gradient tensors are quantized to NVFP4 which enables use of NVFP4 GEMMs in fprop, dgrad, and wgrad. On GB300, peak FP4 throughput is 3$\times$ higher than FP8 throughput~\citep{nvidia2025blackwellultra}. Prior work in NVFP4 pretraining ~\citep{nvidia2025pretraininglargelanguagemodels} simulated NVFP4 numerics through quantize-dequantize functions around BF16 GEMMs. This work uses native NVFP4 GEMMs, leveraging cuBLAS's backend for Transformer Engine.

The NVFP4 format features: fine-grained (16 element) micro-block scaling, block scaling factors with E4M3 format, a second-level FP32 global scale, and the E2M1 element format. We utilize two-dimensional (2D) block scaling for weight quantization, Random Hadamard Transforms (RHTs) on inputs to wgrad, and stochastic rounding on gradients. We kept the last 15\% of the network in high precision to maintain stability. 

Super and Ultra models employ the Latent-MoE architecture and MTP. We kept latent projections in BF16 as its impact to step-time is minimal. We also keep MTP layers in BF16 due to their positioning at the end of the network and to preserve MTP capabilities.

\begin{figure}[]
    \centering
    \includegraphics[width=\textwidth]{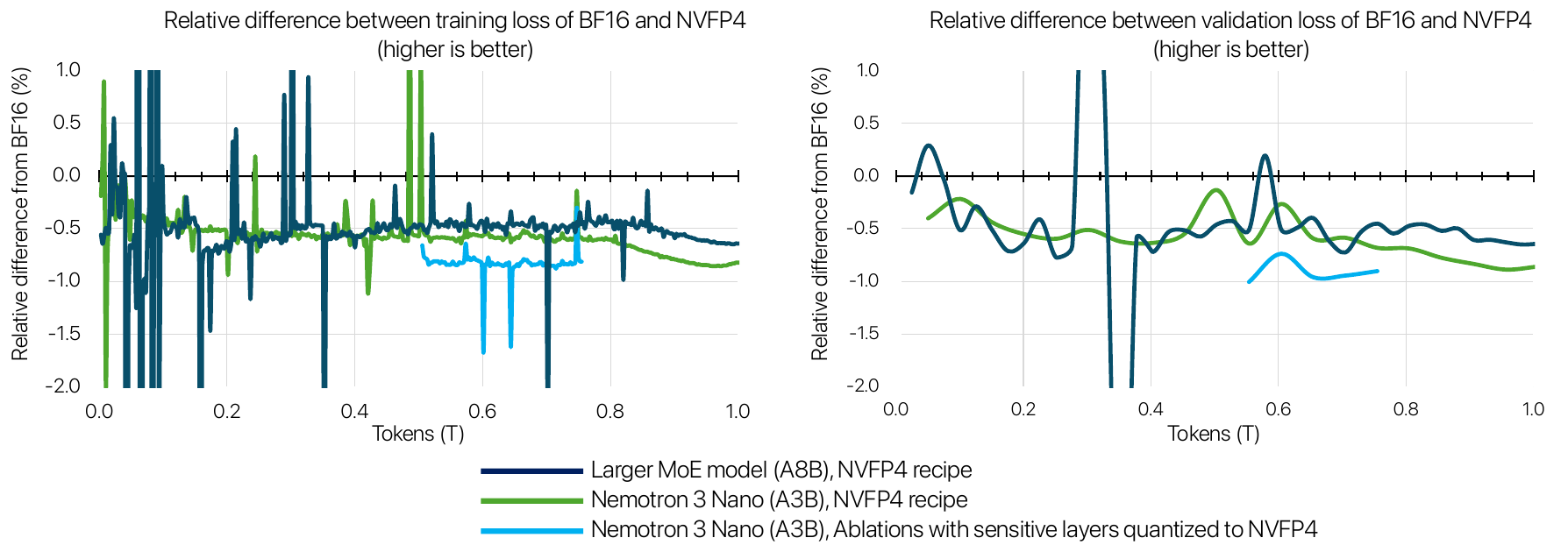}
    \caption{Relative difference in train loss (left) and validation loss (right) between models trained with NVFP4 and BF16, shown at two model scales: Nemotron 3 Nano (A3B) and the larger MoE model (A8B). Loss gaps decrease as model size increases (A3B $\to$ A8B). Recipe ablation on Nemotron 3 Nano started from Nemotron 3 NVFP4 checkpoint at 500B tokens, then quantizes sensitive layers (Mamba Output, QKV, and Attention projections) to NVFP4, highlighting the importance of keeping these layers in high precision.  
    }
    \label{fig:nanov3_loss_diff_nvfp4}
\end{figure}

The Nemotron 3 family of models features a small ratio of attention to Mamba-2 layers, and each attention layer uses GQA with only 2 KV heads. To maintain the fidelity of these few attention layers, we kept the QKV and attention projections in BF16. We observed that Mamba output projection layers have high flushes to zero (up to 40\% on Nano) when quantized to NVFP4. To prevent loss of information, we keep these layers in MXFP8. Figure~\ref{fig:nanov3_loss_diff_nvfp4} shows that combining both modifications resulted in a recipe that improved both train and validation loss (green) compared to keeping these layers in low precision (blue).

\begin{figure}[]
    \centering
    \includegraphics[width=\textwidth]{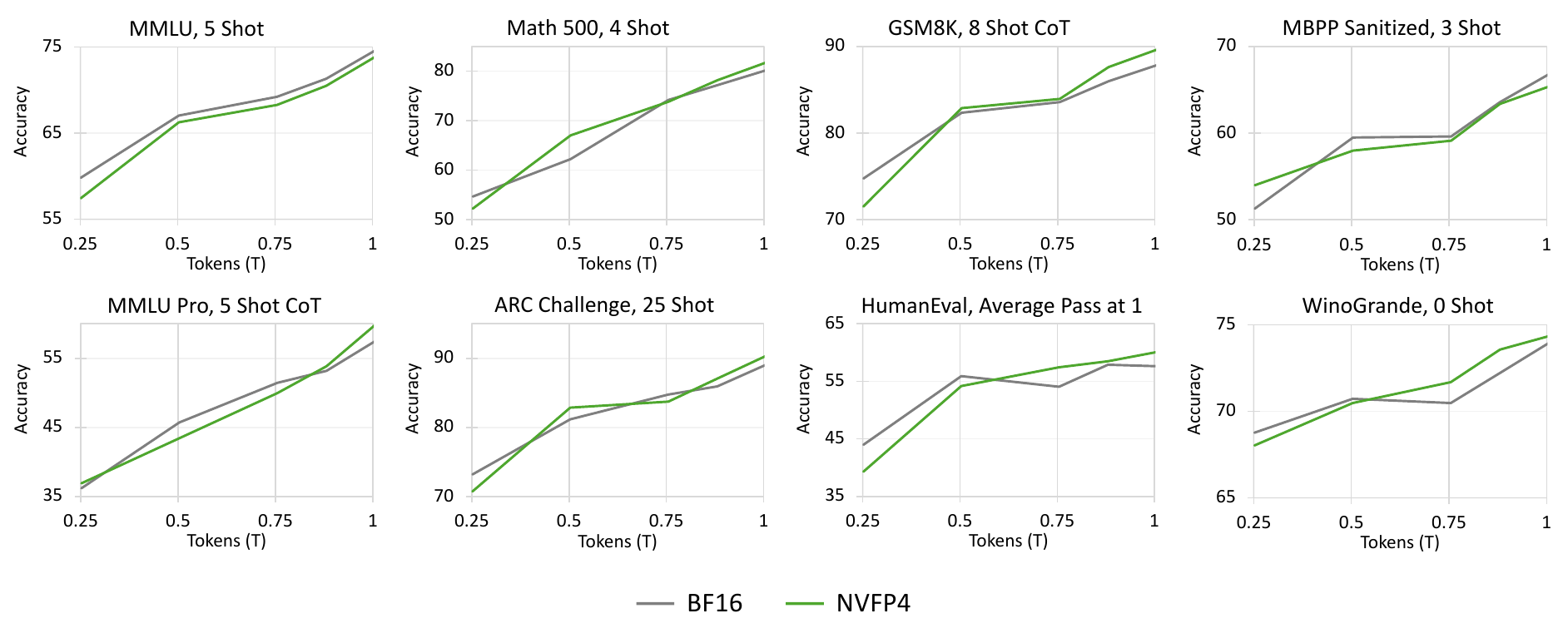}
    \caption{Downstream task evaluations on 8B active MoE model, trained to 1T tokens. NVFP4 accuracy closely follows BF16 trajectories throughout training. Evaluations are performed in BF16.
    }
    \label{fig:evals_nvfp4_8b}
\end{figure}

Figure~\ref{fig:nanov3_loss_diff_nvfp4} also shows relative loss gaps between NVFP4 and BF16. On Nano, we achieve a < 1\% relative difference in loss between NVFP4 vs BF16 (green). The loss gap decreases to < 0.6\% between NVFP4 vs BF16 when trained on a larger MoE model with 8B active parameters (dark blue). Prior art further reinforces these findings that loss gaps induced from quantization decrease as model size increase ~\citep{chen2025scalinglawquantizationawaretraining}. Downstream task evaluations shown in Figure~\ref{fig:evals_nvfp4_8b} are comparable between the A8B model trained in BF16 vs NVFP4. This phenomenon further confirms prior work on Mamba-MLP models where a small loss gap does not lead to degraded evaluation accuracy ~\citep{nvidia2025pretraininglargelanguagemodels}.

\subsection{Long Context}
\label{subsec:lc}

The Nemotron 3 models are designed to support context lengths up to 1M tokens to enable extended multi-turn agentic reasoning. Rotary Position Embeddings (RoPE) are a known hurdle to extending context beyond the training length. Since Mamba layers provide implicit positional information, Nemotron 3 models do not use RoPE in attention layers and therefore do not suffer from out-of-distribution RoPE issues during context extension (a Transformer analog is explored in \cite{puvvada-etal-2025-swan}). For \ourmodel, we included a continued pre-training (CPT) stage at a 512k sequence length, and supervised fine-tuning (SFT) was performed at a 256k sequence length. In addition, we include a long-context environment in the reinforcement learning stage with inputs up to 32k tokens. All three stages included synthetic data designed to support long-range retrieval, multi-hop reasoning, multi-document information aggregation, and related capabilities. In CPT, we did not observe the need to follow a staged increase of training sequence length from 8k to 512k. Further, we observe that the MoE hybrid architecture adopted for the Nemotron 3 models has better context extension capability compared to the dense hybrid architecture used in Nemotron 2 Nano. When continue pre-trained on the same sequence length (512k), the Nemotron 3 Nano base model shows better RULER \citep{hsieh2024ruler} scores compared to the Nemotron 2 Nano 12B base model at a 1M context length (Table \ref{tab:base_ruler}).

\begin{figure}[h]
    \centering
    \includegraphics[width=0.5\linewidth]{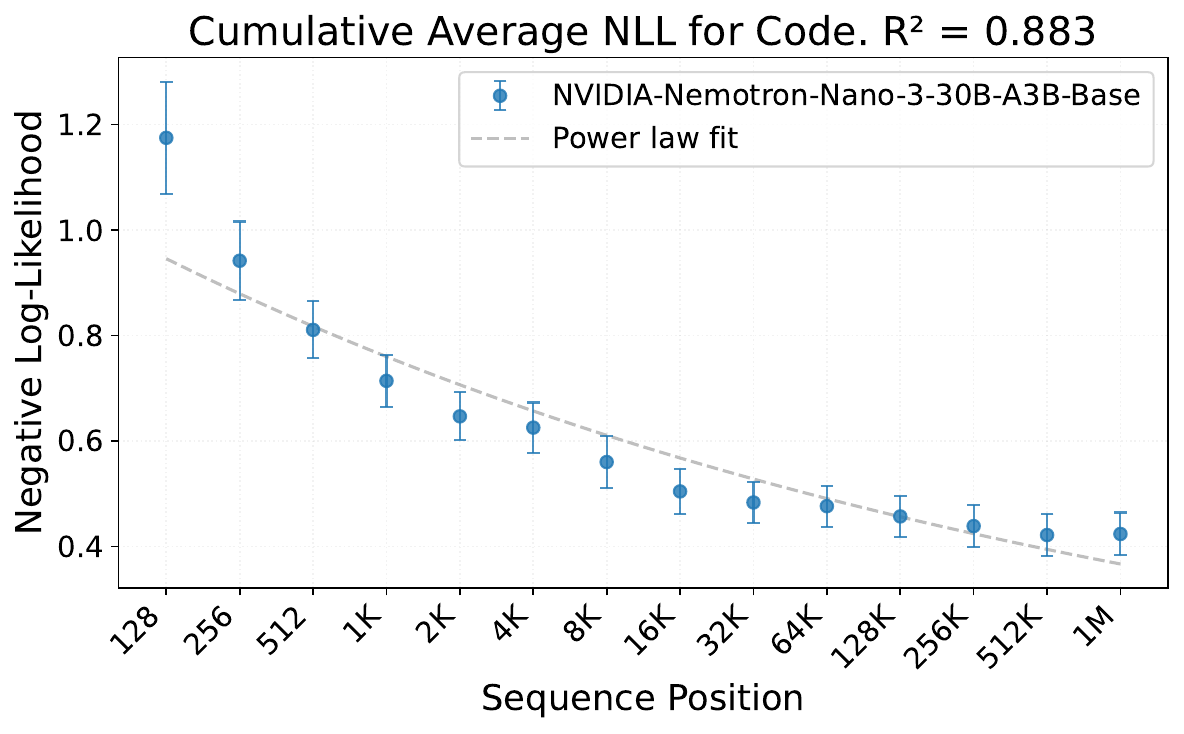}
    \caption{Cumulative average Negative loglikelihood (NLL) as a function of token position in code data. \ourmodel base shows improved predictions upto 1M tokens in code data. 
    }
    \label{fig:nano_v3_base_nll}
\end{figure}

To further assess how well Nemotron 3 Nano leverages very long contexts for next-token prediction, we measure the negative log-likelihood (NLL) of tokens at various positions in held-out sequences. Lower NLL indicates better predictive performance. In a related coherent sequence, tokens appearing later in the context should be easier to predict and therefore exhibit lower NLL. We conduct this analysis on repository-level code sequences larger than 1 million tokens. Figure \ref{fig:nano_v3_base_nll} shows the cumulative average NLL up to each token index for \ourmodel base. We observe that NLL decreases with sequence length, suggesting that the model is able to use long input context up to the tested range.

\begin{table*}[h]
\centering
\begin{tabular}{l|cccc}
\toprule
\textbf{Model} & \textbf{128k} & \textbf{256k} & \textbf{512k} & \textbf{1M} \\
\midrule
Nemotron-Nano-12B-v2-Base & 85.13 & 79.85 & 75.12 & 23.43  \\
\ourbasemodel & 74.48 & 71.67 & 66.02 & \textbf{54.19}  \\
\bottomrule
\end{tabular}
\caption{RULER scores for Nemotron-Nano-12B-v2-Base (Dense  Hybrid) and \ourbasemodel (MoE hybrid) models at different input context lengths. We observe that MoE hybrid model is more robust to length extrapolation than dense Hybrid Model. Nemotron-Nano-12B-v2-Base model shows abrupt dropoff between 512k and 1M, where as \ourbasemodel exhibits graceful degradation. Both models were trained upto 512k sequence length.}
\label{tab:base_ruler}
\end{table*}

\subsection{Multi-environment Reinforcement Learning Post-training}
\label{subsec:multienvRL}
\begin{figure}[]
    \centering
    \includegraphics[width=0.9\textwidth]{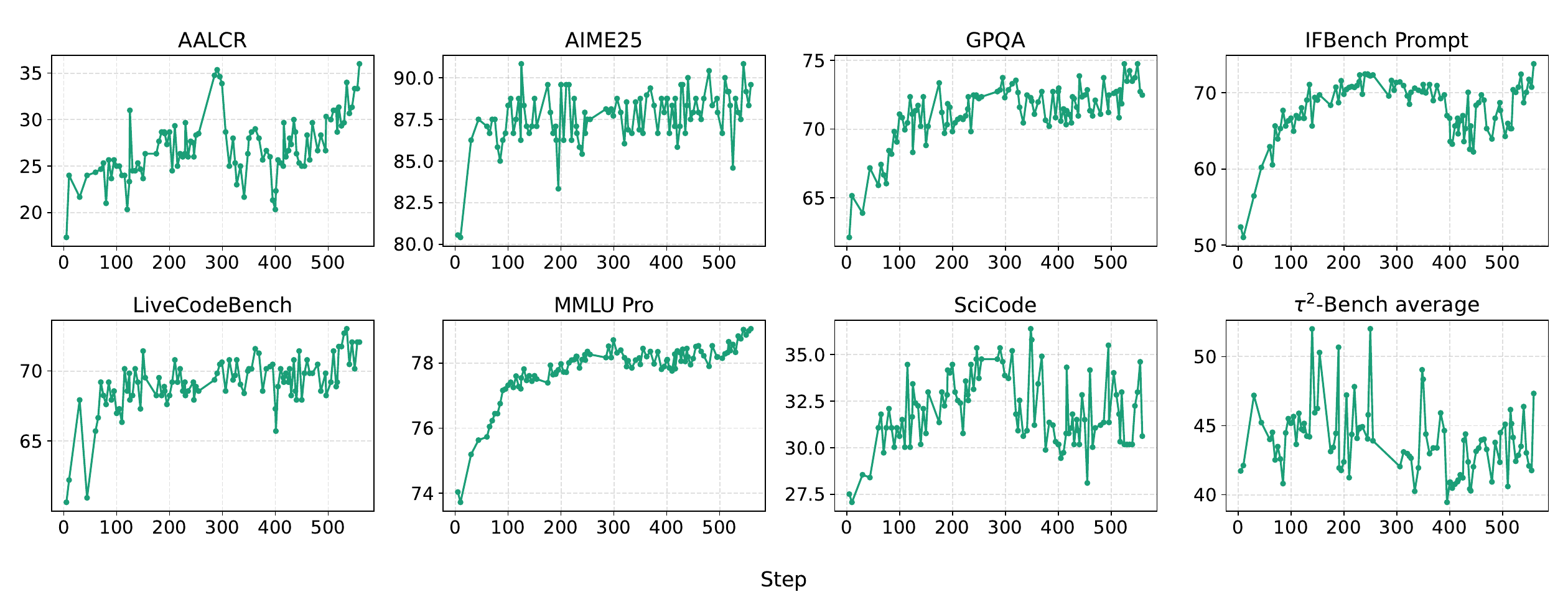}
    \caption{Multi-environment RL training: within a single RL run several different environments corresponding to diverse capabilities are being optimized.
    }
    \label{fig:multi-env-rl}
\end{figure}

The Nemotron 3 models are designed to serve as the foundation for a wide variety of agentic AI applications. To teach Nemotron 3 the capabilities needed to succeed across such a broad range of tasks, we create a diverse set of reinforcement learning (RL) environments, covering mathematical and scientific reasoning, competitive coding, instruction following, software engineering, search, chat, general agentic tool use, long context, and more. Unlike our previous models where we had separate training stages for different tasks, we train the Nemotron 3 models on all of these tasks simultaneously. We find such simultaneous training is more stable, less prone to reward hacking and overall better compared to previous staged approaches \citep{basant2025nvidia}, which often results in degradation of some capabilities \citep{deepseekai2024deepseekv32}. The utility of multi-environment RL can be seen in Figure \ref{fig:multi-env-rl}, where performance on a wide variety of agentic and reasoning benchmarks steadily increases throughout Nemotron 3 Nano RL training.

Large-scale RL across heterogeneous and complex environments requires efficient system design coupled with stable learning algorithms. The Nemotron 3 models are well suited for this setting, as their high inference throughput provides a significant advantage during large-scale rollout generation compared to other open-source models
. To further improve sampling efficiency, we employ an asynchronous RL architecture that decouples training from inference and leverage multi-token prediction to accelerate rollout generation. For stable training, we use GRPO \citep{deepseek-math} with masked importance sampling to account for discrepancies between the training and rollout policies.

Our entire post-training SW stack is open-sourced under Apache 2.0. NeMo-RL \footnote{\url{https://github.com/NVIDIA-NeMo/RL}} implements scalable RL training while NeMo-Gym \footnote{\url{https://github.com/NVIDIA-NeMo/Gym}} provides collection of RL environments.





\subsection{Granular Reasoning Budget Control at Inference Time}
\label{subsec:budgetcontrol}

\begin{figure}[]
    \centering
    \includegraphics[width=0.9\textwidth]{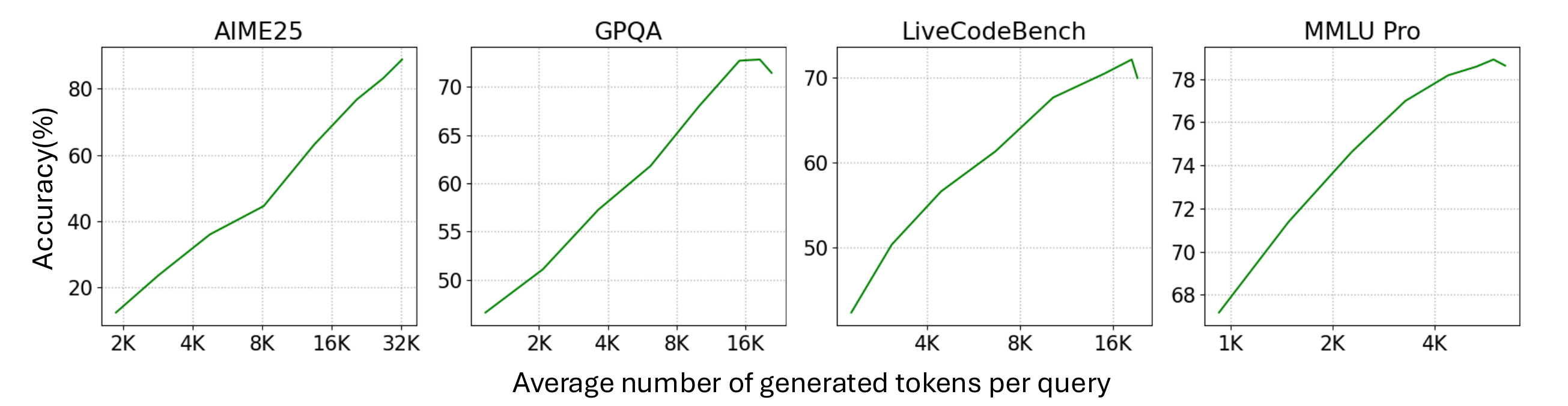}
    \caption{Accuracy-efficiency trade-off with reasoning budget control at inference time.
    }
    \label{fig:budgets}
\end{figure}

Similar to Nemotron 2 Nano~\citep{basant2025nvidia}, the Nemotron 3 models are trained to work with inference-time budget control. Given a user-specified budget on the max number of tokens to use in a thinking trace and when the model reaches the budget, one can append the \texttt{</think>} token to the sequence and let the model continue to generate. The model will generate the response based on the partial thinking trace. Figure~\ref{fig:budgets} illustrates the accuracy-efficiency trade-off curves of Nemotron 3 Nano by varying the token budget, and this provides users fine-grained control in AI applications.

\section{Key Takeaways}
\label{sec:contributions}
In this white paper, we introduced the Nemotron 3 family of models: Nano, Super, and Ultra. Nemotron 3 is the most efficient family of open models with leading accuracy for building high-accuracy agentic AI applications. Nemotron 3 models use a hybrid Mamba-Transformer MoE architecture, are trained using multiple RL environments, offer granular reasoning budget control, and support a context length of up to 1M tokens. Super and Ultra push the improvements further by using LatentMoE and NVFP4 training. Super and Ultra will also ship with MTP layers for enabling fast, low-latency generation. Nemotron 3 models will be open and transparent - we will release model weights, pre- and post-training software, training recipes, and most of the training data. Nemotron 3 Nano is released along with this white paper. Super and Ultra releases will follow in the upcoming months.
\section*{Contributors}

We thank the following people for their invaluable contributions to the Nemotron 3 family of models.

\textbf{Pretraining Data.} Abhinav Khattar, Aleksander Ficek, Alisa Liu, Arham Mehta, Asif Ahamed, Ayush Dattagupta, Benedikt Schifferer, Brandon Norick, Branislav Kisacanin, Dan Su, Dane Corneil, Daria Gitman, Dhruv Nathawani, Dima Rekesh, Divyanshu Kakwani, Edgar Minasyan, Eileen Long, Ellie Evans, Eric Tramel, Evelina Bakhturina, Felipe Soares, Feng Chen, Gantavya Bhatt, George Armstrong, Igor Gitman, Ivan Moshkov, Jane Polak Scowcroft, John Kamalu, Johnny Greco, Joseph Jennings, Jupinder Parmar, Kezhi Kong, Markus Kliegl, Maarten Van Segbroeck, Matvei Novikov, Mehrzad Samadi, Miguel Martinez, Mohammad Shoeybi, Mostofa Patwary, Nabin Mulepati, Oleksii Hrinchuk, Rabeeh Karimi Mahabadi, Rima Shahbazyan, Riyad Islam, Roger Waleffe, Rohit Watve, Sadegh Mahdavi, Sanjeev Satheesh, Sean Narentharen, Shrimai Prabhumoye, Shubham Pachori, Shubham Toshniwal, Shuoyang Ding, Somshubra Majumdar, Stephen Ge, Sumeet Kumar Barua, Suseella Panguluri, Syeda Nahida Akter, Vahid Noorozi, Vitaly Kurin, Vitaly Lavrukhin, Wasi Uddin Ahmad, Wei Du, Wei Ping, Yejin Choi, Yev Meyer, Ying Lin, Zihan Liu

\textbf{Architecture.} Abhinav Khattar, Bita Darvish Rouhani, Deepak Narayanan, Ilya Loshchilov, Jatin Mitra, Joey Guman, Mohammad Shoeybi, Mostofa Patwary, Kezhi Kong, Krishna C. Puvvada, Maor Ashkenazi, Nidhi Bhatia, Pavlo Molchanov, Rabeeh Karimi Mahabadi, Rasoul Shafipour, Ritika Borkar, Roger Waleffe, Ryan Prenger, Sanjeev Satheesh, Venmugil Elango, Yonggan Fu

\textbf{Pretraining Software.} Aarti Basant, Ashwath Aithal, Abhinav Khattar, Deepak Narayanan, Duncan Riach, Eric Harper, Hexin Wang, Jared Casper, Jimmy Zhang, Kezhi Kong, Mike Chrzanowski, Nima Tajbakhsh, Pranav Prashant Thombre, Roger Waleffe, Russell J. Hewett, Seonmyeong Bak, Shiqing Fan, Vijay Korthikanti, Xiaowei Ren, Yashaswi Karnati, Zijie Yan

\textbf{Pretraining.} Abhinav Khattar, Brandon Norick, Dan Su, Eric Tramel, Deepak Narayanan, John Kamalu, Joseph Jennings, Jupinder Parmar, Markus Kliegl, Miguel Martinez, Mohammad Shoeybi, Mostofa Patwary, Kezhi Kong, Kevin Shih, Rabeeh Karimi Mahabadi, Roger Waleffe, Ryan Prenger, Shrimai Prabhumoye, Sanjeev Satheesh, Syeda Nahida Akter, Ying Lin 

\textbf{NVFP4.} Abhinav Khattar, Aditya Vavre, Anjulie Agrusa, Ankur Verma, Asit Mishra, Asma Kuriparambil Thekkumpate, Ben Lanir, Bita Darvish Rouhani, Bryan Catanzaro, Carlo del Mundo, Cyril Meurillon, Daniel Lo, Daria Levy, Darko Stosic, Deepak Narayanan, Dusan Stosic, Eric Chung, Eric Tramel, Evgeny Tsykunov, Frank Sun, Herbert Hum, Jimmy Zhang, Jinhang Choi, Jining Huang, Keith Wyss, Kirthi Shankar, Lizzie Wei, Mahdi Nazemi, Matt Kulka, Michael Andersch, Mikail Khona, Mike Chrzanowski, Minseok Lee, Mohammad Shoeybi, Mostofa Patwary, Nima Tajbakhsh, Nishant Sharma, Pasha Shamis, Paul Gibbons, Przemek Tredak, Qiyu Wan, Rabeeh Karimi Mahabadi, Rachit Garg, Robert Hesse, Roger Waleffe, Russell Hewett, Sangkug Lim, Sanjeev Satheesh, Stas Sergienko, Tim Moon, Victor Cui, Vinay Rao, Xiaowei Ren, Yigong Qin, Zhongbo Zhu

\textbf{Long Context.} Boris Ginsburg, Cheng-Ping Hsieh, Dan Su, Dima Rekesh, Faisal Ladhak, Fei Jia, John Kamalu, Kezhi Kong, Krishna C. Puvvada, Markus Kliegl, Mostofa Patwary, Roger Waleffe, Samuel Kriman, Sanjeev Satheesh, Shantanu Acharya, Simeng Sun, Ushnish De

\textbf{Posttraining Software.} Adi Renduchintala, Alexander Bukharin, Ali Taghibakhshi, Banghua Zhu, Brian Yu, Duncan Riach, Frankie Siino, Gerald Shen, Jiaqi Zeng, Kezhi Kong, Li Ding, Luis Vega, Maanu Grover, Marc Romeijn, Parth Chadha, Peter Jin, Soumye Singhal, Terry Kong, Tugrul Konuk, Yi-Fu Wu, Yubo Gao

\textbf{Posttraining.} Abhibha Gupta, Adi Renduchintala, Akanksha Shukla, Aleksander Ficek, Alexander Bukharin, Ameya Sunil Mahabaleshwarkar, Banghua Zhu, Besmira Nushi, Branislav Kisacanin, Cheng-Ping Hsieh, Charles Wang, Damon Mosk-Aoyama, Daria Gitman, Dhruv Nathawani, Dima Rekesh, Edgar Minasyan, Edward Lin, Evelina Bakhturina, Fei Jia, Felipe Soares, Feng Chen, George Armstrong, Grigor Nalbandyan, Haifeng Qian, Hayley Ross, Igor Gitman, Ivan Moshkov, Jeffrey Glick, Jiaqi Zeng, Jian Zhang, Jie Lou, Julien Veron Vialard, Junkeun Yi, Katherine Luna, Khushi Bhardwaj, Krishna C. Puvvada, Luis Vega, Makesh Narsimhan Sreedhar, Matvei Novikov, Mehrzad Samadi, Mengru Wang, Michael Evans, Nikolai Ludwig, Oleksii Hrinchuk, Oleksii Kuchaiev, Olivier Delalleau, Ouye Xie, Peter Jin, Pritam Gundecha, Prasoon Varshney, Rima Shahbazyan, Ritu Gala, Sadegh Mahdavi, Sahil Modi, Sanjay Kariyappa, Sean Narenthiran, Shantanu Acharya, Shubham Toshniwal, Shuoyang Ding, Somshubra Majumdar, Soumye Singhal, Stephen Ge, Sugam Dipak Devare, Suseella Panguluri, Tugrul Konuk, Vahid Noroozi, Venkat Srinivasan, Vitaly Lavrukhin, Wasi Uddin Ahmad, Wei Du, Yev Meyer, Yian Zhang, Yoshi Suhara

\textbf{Evaluation, Safety and Release.} Aaron Grattafiori, Barnaby Simkin, Besmira Nushi, Bilal Kartal, Christopher Parisien, Daniel Rohrer, David Mosallanezhad, Eileen Peters Long, Erick Galinkin, Fay Wang, Ferenc Galko, Gorkem Batmaz, Jane Polak Scowcroft, Katherine Luna, Khushi Bhardwaj, Leon Derczynski, Michael Boone, Michael Evans, Piotr Januszewski, Rich Harang, Rishabh Garg, Riyad Islam, Sanjay Kariyappa, Sanjeev Satheesh, Shaona Ghosh, Wojciech Prazuch, Yoshi Subara, Zhen Dong, Zijia Chen

\textbf{Infrastructure.} Aaron Blakeman, Anubhav Mandarwal, Alex Kondratenko, Aleksandr Shaposhnikov, Ashwin Poojary, Brandon Soubasis, Collin Neale, Dong Ahn, Evan Briones, Gargi Prasad, Harsh Sharma, Herman Sahota, Himanshu Soni, Jining Huang, Kumar Anik, Maer Rodrigues de Melo, Nikhil Jukar, Pasha Shamis, Rick Izzo, Ruoxi Zhang, Satish Pasumarthi, Sergey Kashirsky, Shelby Thomas, Stefania Alborghetti

\textbf{Quantization.} Aditya Vavre, Akhiad Bercovich, Ameya Sunil Mahabaleshwarkar, Amnon Geifman, Asma Kuriparambil Thekkumpate, Ben Lanir, Bilal Kartal, Chenhan Yu, Daniel Afrimi, Darko Stosic, Dusan Stosic, Ganesh Ajjanagadde, Huizi Mao, Ido Shahaf, Jenny Chen, Kai Xu, Nave Assaf, Omer Ullman Argov, Ran Zilberstein, Sharath Turuvekere Sreenivas, Sweta Priyadarshi, Tijmen Blankevoort, Tomer Asida, Yoshi Suhara,  Zach Moshe, Zijia Chen

\textbf{Inference.} Amir Klein, Amit Zuker, Chenghao Zhang, Daniel Afrimi, Daniel Serebrenik, Gal Hubara Agam, Helen Ngo, Joyjit Daw, Kan Zhu, Keshav Santhanam, Lawrence McAfee, Lucas Liebenwein, Luis Vega, Nave Assaf, Neta Zmora, Netanel Haber, Omer Ullman Argov, Peter Dykas, Pranav Prashant Thombre, Ran Zilberstein, Roi Koren, Shahar Mor, Shanmugam Ramasamy, Siddharth Singh, Suyog Gupta, Teodor-Dumitru Ene, Tomer Asida, Tomer Bar Natan, Vijay Korthikanti, Wanli Jiang, William Zhang, Yashaswi Karnati

\textbf{Compression.} Akhiad Bercovich, Ali Taghibakhshi, Amnon Geifman, Elad Segal, Ido Galil, Ido Shahaf, Itamar Schen, Itay Levy, Izik Golan, Marcin Chochowski, Mohammad Dabbah, Mostofa Patwary, Najeeb Nabwani, Nave Assaf, Nir Ailon, Omri Puny, Oren Tropp, Pavlo Molchanov, Ran El-Yaniv, Ran Zilberstein, Saurav Muralidharan, Sharath Turuvekere Sreenivas, Tomer Asida, Tomer Ronen, Vladimir Anisimov, Yonatan Geifman, Zach Moshe

\textbf{Deployment.} Alexandre Milesi, Anahita Bhiwandiwalla, Huy C Nguyen, Huy Q Nguyen, Izzy Putterman, Manoj Kilaru, Maryam Moosaei, Pawel Morkisz, Tan Bui, Thanh Do

\textbf{Legal and Compliance.} Barnaby Simkin, Chantal Hwang, Chetan Mungekar, Dina Yared, Hiren Upadhyay, Iain Cunningham, Katherine Cheung, Laya Sleiman, Meredith Price, Michael Boone, Nikki Pope, Saori Kaji

\textbf{Marketing.} Amelia Barton, Chintan Patel, Erik Pounds, Mark Cai, Natalie Hereth, Nicola Sessions, Nirmal Juluru, Shreya Gopal, Will Jennings

\textbf{Project Management.} Amy Shen, Ann Guan, Bardiya Sadeghi, Daria Levy, Elena Lantz, Elliott Ning, Krzysztof Pawelec, Melissa Corpuz, Negar Habibi, Pinky Xu, Qing Miao, Ryan Timbrook, Seth Poulos, Smita Ithape, Twinkle Vashishth

\textbf{Product.}  Chris Alexiuk, Ellie Evans, Jane Polak Scowcroft, Jesse Oliver, Joey Conway, Tom Balough, Udi Karpas, Wenfei Zhou

\textbf{Leadership.} Andrew Tao, Bita Darvish Rouhani, Boris Ginsburg, Bryan Catanzaro, Carlo del Mundo, Eileen Long, Eric Chung, Jane Polak Scowcroft, Jan Kautz, Jian Zhang, Joey Conway, Jonathan Cohen, Kari Briski, Mohammad Shoeybi, Mostofa Patwary, Oleksii Kuchaiev, Oluwatobi Olabiyi, Pavlo Molchanov, Ran El-Yaniv, Ran Zilberstein, Yonatan Geifman, Yejin Choi

\newpage

\bibliography{references}
\bibliographystyle{references}

\end{document}